\documentclass[10pt,twocolumn,letterpaper]{article}

\usepackage{iccv}              
\usepackage[accsupp]{axessibility}
\usepackage{multirow} 
\usepackage{multicol}
\usepackage{amsthm}
\usepackage[dvipsnames,table]{xcolor}

\usepackage{colortbl}

\newtheorem{definition}{Definition}[section] 
\newtheorem{proposition}{Proposition}[section]
\definecolor{iccvblue}{rgb}{0.21,0.49,0.74}
\usepackage[pagebackref,breaklinks,colorlinks,allcolors=iccvblue]{hyperref}

\def\paperID{11176} 
\def\confName{ICCV}
\def\confYear{2025}

\title{Integrating Biological Knowledge for Robust Microscopy Image Profiling \\ on \textit{De Novo} Cell Lines}

\author{Jiayuan Chen\thanks{Equal contribution.}\and Thai-Hoang Pham\footnotemark[1]\and Yuanlong Wang \and Ping Zhang\thanks{Corresponding author.}\\
The Ohio State University\\
{\tt\small \{chen.12930,pham.375,wang.16050,zhang.10631\}@osu.edu}
}

\begin{document}
\maketitle
\begin{abstract}
High-throughput screening techniques, such as microscopy imaging of cellular responses to genetic and chemical perturbations, play a crucial role in drug discovery and biomedical research. However, robust perturbation screening for \textit{de novo} cell lines remains challenging due to the significant morphological and biological heterogeneity across cell lines. To address this, we propose a novel framework that integrates external biological knowledge into existing pretraining strategies to enhance microscopy image profiling models. Our approach explicitly disentangles perturbation-specific and cell line-specific representations using external biological information. Specifically, we construct a knowledge graph leveraging protein interaction data from STRING and Hetionet databases to guide models toward perturbation-specific features during pretraining. Additionally, we incorporate transcriptomic features from single-cell foundation models to capture cell line-specific representations. By learning these disentangled features, our method improves the generalization of imaging models to \textit{de novo} cell lines. We evaluate our framework on the RxRx database through one-shot fine-tuning on an RxRx1 cell line and few-shot fine-tuning on cell lines from the RxRx19a dataset. Experimental results demonstrate that our method enhances microscopy image profiling for \textit{de novo} cell lines, highlighting its effectiveness in real-world phenotype-based drug discovery applications.
\end{abstract}    
\section{Introduction}
\label{sec:intro}
Cellular responses to genetic and chemical perturbations support drug discovery and biomedical research by providing critical insights into disease mechanisms and therapeutic treatments \cite{replogle2022mapping,way2022morphology,pham2021deep}. In particular, fluorescence microscopy image analysis—the process of extracting meaningful information about cellular responses from fluorescence images captured by a microscopy—facilitates the identification of novel drug candidates and the prediction of treatment outcomes, extending beyond traditional target-based drug discovery approaches \cite{pun2023ai,lee2019deepconv}. With advancements in high-throughput screening (HTS) techniques, we can now capture large-scale microscopy data about cellular morphology \cite{rxrx1,chandrasekaran2024three,bray2016cell}. On the other hand, machine learning (ML) techniques have evolved rapidly to enable more accurate and scalable image analyses \cite{carpenter2006cellprofiler,dosovitskiy2020image, he2016deep}. Together, these advancements are transforming phenotype-based drug discovery by enabling more comprehensive, data-driven approaches to understand cellular behavior and identifying promising therapeutic treatments \cite{espinosa2016duality,mcdonald2003imaging,cloome}.
\begin{figure}
    \centering
\includegraphics[width=0.92\linewidth]{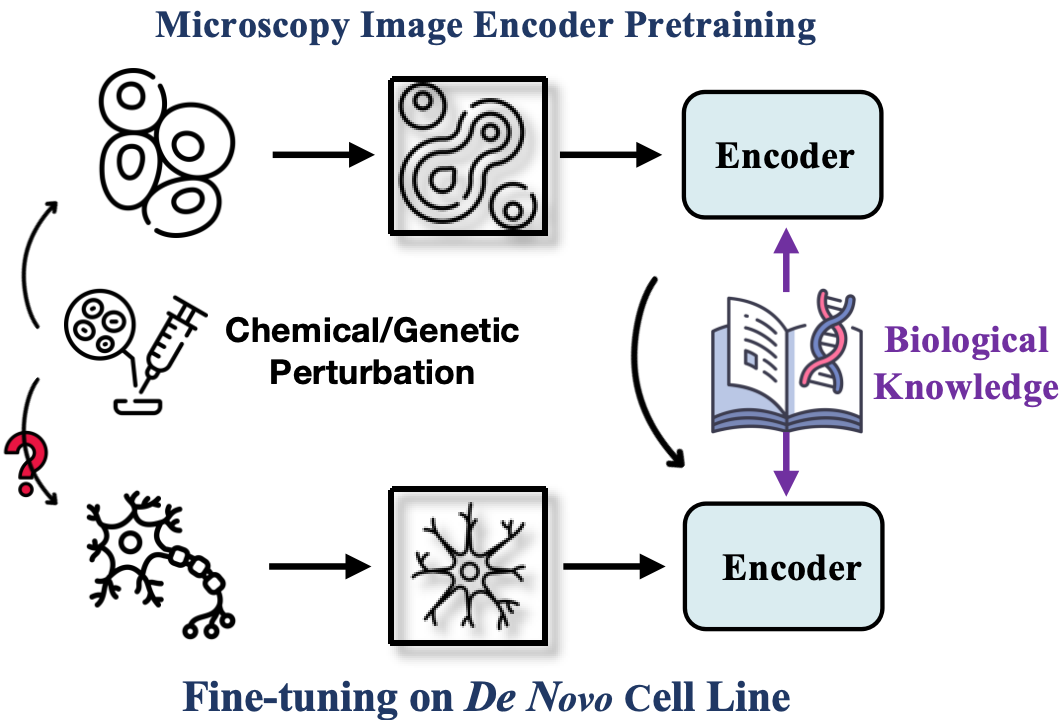}
    \caption{Microscopy image perturbation screening and transfer to de novo cell lines. Our method integrates biological priors, including protein interactions and cell line gene expression, to improve model robustness.}
    \label{fig:1}
\end{figure}
With the vast availability of microscopy data capturing cellular responses, pretraining has emerged as a dominant approach in deep learning for microscopy image analysis. By training on large-scale microscopy datasets before fine-tuning on specific tasks, pretrained models can leverage learned generalizable features to enhance performance in specialized applications, such as perturbation screening \cite{rxrx1,moshkov2024learning} and mechanism of action identification \cite{cloome,kraus2024masked}. While pre-training strategies have successfully expanded microscopy image analysis to \textit{de novo} genetics and chemical perturbations, existing methods are not designed to handle \textit{de novo} cell lines. Given the inherent challenges of \textit{in vitro} perturbation screening—such as high costs and the labor-intensive nature of data collection \cite{macarron2011impact,huang2022artificial}—developing computational models which are robust to cell heterogeneity is crucial. In this work, we address this limitation by focusing on conducting perturbation screening for \textit{de novo} cell lines, assessing the generalization capabilities of microscopy image analysis across previously unseen cell lines.

Existing pretraining strategies for fluorescence microscopy images, such as weakly supervised learning with chemical and genetic perturbation labels \cite{moshkov2024learning} and self-supervised methods like masked autoencoders (MAE) \cite{kraus2024masked} and self-distillation (DINO) \cite{sivanandan2023pooled}, have demonstrated promising results in perturbation screening for \textit{de novo} perturbations. However, leveraging these pretrained models for perturbation screening in \textit{de novo} cell lines remains challenging due to the significant morphological and biological heterogeneity across cell types. To enhance model generalization to \textit{de novo} cell lines, we hypothesize that the model must explicitly capture cell line-specific and perturbation-specific information during pretraining. To achieve this, we propose integrating external biological knowledge to guide the model in learning disentangled representations, as illustrated in \autoref{fig:1}. First, we incorporate transcriptomic information from diverse cell lines using gene-gene relationships aggregated from multiple databases, including STRING \cite{szklarczyk2023string} and Hetionet \cite{himmelstein2017systematic}, which provide a comprehensive resource of protein-protein interactions (PPIs) across multiple organisms. By constructing a knowledge graph where microscopy images serve as nodes and PPI relationships define edges, our approach helps the model focus on perturbation-specific patterns during pretraining. Second, to ensure explicit cell-specific representation learning, we leverage a single-cell foundation model \cite{cui2024scgpt,hao2024large,lopez2018deep} to encode RNA-seq data from each cell line and integrate it into the microscopy image analysis. By fusing perturbation-specific features extracted from microscopy images with cell line-specific information derived from the single-cell foundation model, our approach enables the model to learn robust features, improving its generalization to \textit{de novo} cell lines and facilitating accurate, biologically meaningful perturbation screening.


We evaluate the effectiveness of our proposed method for perturbation screening on two RxRx\footnote{https://www.rxrx.ai/datasets} datasets: RxRx1 and RxRx19a. Our focus is on the \textit{de novo} cell line setting, where the cell lines in the training and testing sets do not overlap. Specifically, we pretrain our model on one subset of RxRx1 and evaluate it on a distinct partition of that dataset, as well as on RxRx19a.  To ensure no overlap, we split the RxRx1 data by cell lines, guaranteeing that those in the testing set are absent from the training set. For evaluation on the RxRx1 dataset, the model must learn robust imaging features to generalize its predictions to unseen cell lines. The evaluation on the RxRx19a dataset presents an even greater challenge, as the model must handle both the \textit{de novo} cell line scenario and the heterogeneity in imaging protocols between the two datasets.

In summary, our contributions include:

\begin{itemize}[leftmargin=*]
\item We introduce a realistic but more challenging setting for microscopy image profiling and establish a benchmark for pretraining methods using RxRx database.

\item We propose a novel approach that integrates biological knowledge to guide microscopy image profiling models to learn robust imaging features that help them to generalize their predictions performance to \textit{de novo} cell lines.

\item  Through extensive experiments\footnote{https://github.com/The-Real-JerryChen/BioMicroscopyProfiler}, we demonstrate that our method significantly improves existing pretraining strategies for perturbation classification on \textit{de novo} cell lines, underscoring its potential impact on drug discovery.
\end{itemize}

\begin{figure*}
    \centering
    \includegraphics[width=0.9\linewidth]{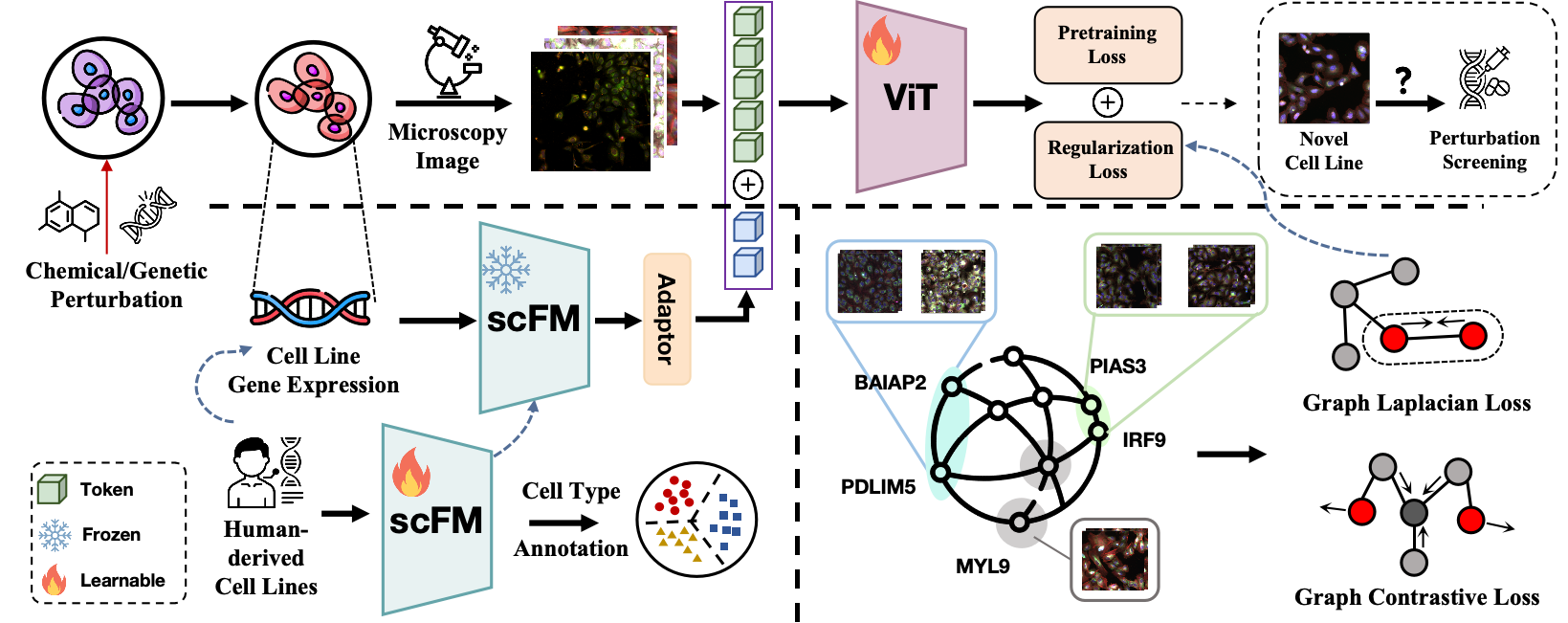}
    \caption{Overview of our proposed method. The upper section illustrates the baseline framework for perturbation screening based on microscopy images. The bottom right section details the construction of the perturbation relational graph and the two graph regularization losses based on it, where each node in the graph represents a perturbation gene. The bottom left part demonstrates how we learn cell line-specific features and then incorporate them into the image encoder.}
    \label{fig:2}
\end{figure*}
\section{Related Work}
\subsection{Cellular Response Data}
Cellular response data serve as a critical resource in drug discovery and have garnered significant attention in recent years \cite{zhang2024single,wang2016drug,haghighi2022high}. Advances in HTS have enabled the collection of massive datasets that capture diverse cellular responses to various perturbations. For example, cellular gene expression data, such as those generated by the L1000 platform, provide bulk measurements that reflect average transcriptional changes across cell populations under different treatments \cite{subramanian2017next}. More recently, image-based assays have emerged as a powerful approach to capture phenotypic responses at a high resolution. Techniques like Cell Painting generate fluorescent microscopy images that reveal a wealth of morphological information, including changes in cell shape, texture, and subcellular organization \cite{chandrasekaran2024three,bray2016cell}. These images document cellular responses to various perturbations, ranging from genetic modifications to chemical treatments. The integration of these cellular datasets has opened new avenues in drug discovery, facilitating applications such as target identification \cite{akbarzadeh2022morphological,kraus2024masked}, mechanism-of-action studies \cite{harrison2023evaluating,liu2025learning}, and the development of predictive models for drug efficacy and toxicity \cite{way2021predicting, fredin2024cell}. 
\subsection{Microscopy Image Representation Learning}
Microscopy image representation learning is a critical step in phenotype-based drug discovery, as it enables the extraction of meaningful features from perturbation images that reflect underlying cellular responses \cite{celik2022biological,tang2024morphological}. Traditional methods, such as CellProfiler \cite{mcquin2018cellprofiler}, rely on classical image processing techniques. In recent years, deep learning-based representation learning has emerged as a promising alternative. One notable approach employs weakly supervised learning with a causal interpretation, leveraging a large, diverse multi-study dataset to disentangle confounding factors from phenotypic features \cite{moshkov2024learning}. In parallel, self-supervised learning techniques popular in computer vision, for example contrastive learning \cite{chen2020simple}, masked autoencoding \cite{he2022masked}, and self-distillation \cite{caron2021emerging}, have been successfully adapted to microscopy image analysis, enabling robust feature extraction from large-scale unlabeled datasets. Additionally, several multimodal approaches have been proposed \cite{liu2025learning,cloome}; for example, \citet{fradkin2025molecules} propose learning a joint latent space that aligns chemical structure and cellular image, overcoming challenges such as batch effects and inactive perturbations. Another line of work in microscopy image analysis has focused on image generation \cite{palma2025predicting,bigverdi2024gene}. For example, researchers have applied generative adversarial networks and flow-based models to generate images that simulate novel perturbations.

\section{Methodology}
\subsection{Problem Setting}
In this study, we focus on microscopy image analysis on \textit{de novo} cell line setting, where we aim to pretrain image encoding models such as Vision Transformer (ViT) \cite{dosovitskiy2020image} on cellular images associated with a set of training cell lines, denoted as $\mathcal{C}^{tr}$, and evaluate their performance on a distinct set of testing cell lines, $\mathcal{C}^{te}$, ensuring that $\mathcal{C}^{tr} \cap \mathcal{C}^{te} = \varnothing$. This setup allows us to assess the generalization capabilities of the pretrained models when applied to unseen cell lines. We specifically focus on the task of perturbation screening, where the objective is to predict perturbations based on corresponding microscopy images, a critical application in pharmaceutical research \cite{adamson2016multiplexed,liberali2015single}.

\subsection{Proposed Method}
\subsubsection{Overview}
Our proposed method enhances model pretraining by incorporating biological knowledge, enabling the learning of robust visual features from cellular microscopy images while seamlessly integrating with existing pretraining approaches such as MAE \cite{he2022masked} and BYOL \cite{grill2020bootstrap}. Specifically, we leverage gene-gene relationships from external databases to construct a knowledge graph, guiding the model to capture perturbation-specific information. Simultaneously, we introduce cell line-specific transcriptomic tokens, derived from RNA-seq data using a single-cell foundation model \cite{cui2024scgpt,lopez2018deep}, to encode cell line-specific characteristics. These biologically informed components enable a more structured and meaningful feature representation, ultimately improving perturbation screening in \textit{de novo} cell line setting. In the following sections, we provide detailed descriptions of these two key components.

\subsubsection{Learning perturbation-specific representation}
Ideally, a vision encoding model should learn to extract perturbation-specific information from microscopy images to accurately predict the perturbation that induces a given cell morphology. However, due to the limited number of cell lines in the training set, the model may instead rely on spurious features—cell line-specific information that correlates with the labels in training data but lacks a true causal relationship with perturbation effects \cite{wang2023removing}. This reliance on spurious correlations hampers the model’s generalization, leading to performance degradation when applied to \textit{de novo} cell lines. To address this issue, we propose incorporating external biological knowledge during model pretraining to explicitly guide the vision encoding model toward perturbation-specific features. Specifically, we construct a perturbation relational graph, where each node represents a genetic or chemical perturbation from cellular microscopy imaging datasets, and edges between nodes encode similarity weights. The motivation behind this graph is to enable the model to leverage relationships between similar and dissimilar perturbations, improving predictive accuracy while ensuring robustness across different cell lines. Additionally, this biological knowledge is inherently transferable, facilitating generalization to \textit{de novo} cell lines. The edge weights between perturbation nodes are derived from well-established biological resources such as STRING \cite{szklarczyk2023string} and Hetionet \cite{himmelstein2017systematic}, which capture diverse biological relationships, including protein-protein interactions, co-expression patterns, and shared functional pathways. In our constructed graph, each perturbation node is assigned a feature vector extracted from the corresponding cellular image, providing a structured and biologically meaningful representation. A formal definition of the perturbation relational graph follows, with further details on biological data sources and graph construction described in Section \ref{setup}.
\begin{definition}[Perturbation Relational Graph]
The perturbation relational graph \(\mathcal{G} = (\mathcal{V}, \mathcal{E}, W, \psi)\) is a structured representation of perturbations in cellular microscopy datasets, designed to capture relationships between different perturbations. Here, \(\mathcal{V}\) denotes the set of perturbations extracted from microscopy images, while \(\mathcal{E}\) represents the set of edges that connect related perturbations. The set of edge weights, represented by \(W\), quantify the intrinsic similarity between perturbations based on biological knowledge. Additionally, \(\psi: \mathcal{V} \to \mathbb{R}^d\) is a mapping function that assigns each perturbation \(v \in \mathcal{V}\) a feature vector \(f_v\), extracted from its corresponding microscopy images.
\end{definition}

To guide the vision encoding model in capturing perturbation-specific signals, we introduce two regularization techniques applied to the extracted visual features: \textbf{graph Laplacian regularization} and \textbf{graph node contrastive learning}. Let \( F \in \mathbb{R}^{|\mathcal{V}| \times d} \) represent the matrix of node features, where each row \( f_{v_i} \in \mathbb{R}^d \) corresponds to the feature representation of perturbation \( v_i \in \mathcal{V} \). We define the degree matrix \( D \in \mathbb{R}^{|\mathcal{V}| \times |\mathcal{V}|} \) with diagonal entries \( D_{ii} = \sum_{j \in \mathcal{V}} w_{ij} \), where \( w_{ij} \) denotes the edge weight between nodes \( v_i \) and \( v_j \). The Laplacian regularization loss is formulated as:
\[
\mathcal{L}_{\text{lap}} = \operatorname{tr}\left(F^\top (D - W) F\right),
\]
where \( \operatorname{tr} \) denotes the trace operator. This loss enforces feature similarity between strongly connected perturbations (i.e., those with larger edge weights), ensuring that the learned embeddings reflect underlying perturbation relationships. In parallel with the graph Laplacian regularization, we employ a graph node contrastive loss on the perturbation relational graph to enhance perturbation-specific representations. For each perturbation \( v_i \), we define its neighborhood \( \mathcal{N}(i) \) as the set of perturbations with stronger edge weights relative to \( v_i \), while treating all other perturbations as negative examples in a contrastive learning framework. The contrastive learning loss is formulated as:
\[
\mathcal{L}_{\text{con}} = \frac{1}{|\mathcal{N}(i)|} \sum_{v_j \in \mathcal{N}(i)} \log \frac{\exp\big(\operatorname{sim}(f_{v_i}, f_{v_j})/\tau\big)}{\sum_{k \in \mathcal{V}} \exp\big(\operatorname{sim}(f_{v_i}, f_{v_k})/\tau\big)}
\]
where \( \operatorname{sim}(\cdot,\cdot) \) represents a similarity measure (e.g., cosine similarity), and \( \tau \) is a temperature hyperparameter that controls the sharpness of similarity scores. This loss encourages the model to bring feature representations of biologically related perturbations closer together in the latent space while pushing apart those of unrelated perturbations. In our study, we experiment with both graph Laplacian regularization and contrastive learning to assess their ability to guide the vision encoding model toward perturbation-specific information. Since graph characteristics such as density and connectivity can vary across datasets, the relative effectiveness of these regularization techniques may depend on dataset-specific properties. We further analyze these effects in Section \ref{ab:loss}. In addition, we demonstrate that the regularization losses (i.e., $\mathcal{L}_{\text{lap}}$ and $\mathcal{L}_{\text{con}}$) defined on the perturbation relational graph implicitly minimizes the intra-class distances within the same perturbation category, thereby facilitating more effective perturbation screening.
\begin{proposition}
\label{proposition}
Let $\phi: \mathcal{X} \to \mathbb{R}^d$ be a vision encoding model that maps each cellular microscopy image $x \in \mathcal{X}$ to a feature vector. For a given perturbation $v \in \mathcal{V}$, we denote $\mathcal{X}_v \subset \mathcal{X}$ as the set of replicate images corresponding to $v$, and define the perturbation node feature as the averaged visual features over the images: $
f_v = \frac{1}{|\mathcal{X}_v|} \sum_{x\in \mathcal{X}_v} \phi(x)
$. Given a perturbation relational graph where each node $v \in \mathcal{V}$ is assigned the feature $\psi(v) := f_v$, then minimizing the graph regularization loss implicitly minimizes the intra-class distance 
$\frac{2}{|\mathcal{X}_v|\times(|\mathcal{X}_v|-1)}\sum_{x, x' \in \mathcal{X}_v} \|\phi(x) - \phi(x')\|$
for each perturbation $v$. 
\end{proposition}
The overall objective function for the pretraining stage is formulated as a weighted combination of the standard pretraining loss—specific to the chosen pretraining method—and the graph regularization loss, either Laplacian regularization (\(\mathcal{L}_{\text{lap}}\)) or contrastive learning loss (\(\mathcal{L}_{\text{con}}\)). This combined objective ensures that the vision encoding model not only learns rich visual features from microscopy images but also aligns these features with biologically meaningful perturbation relationships. By incorporating graph-based regularization, the model is guided to focus on perturbation-induced cellular responses rather than spurious correlations, ultimately improving generalization to \textit{de novo} cell lines. In Section \ref{baseline}, we provide a detailed explanation of how the graph regularization loss is integrated with different pretraining methods.


\subsubsection{Learning cell line-specific representation}
Cell lines inherently exhibit substantial morphological and transcriptomic heterogeneity, manifesting in differences in cell shape, size, nucleus-to-cytoplasm ratio, and the spatial organization of intracellular organelles \cite{marklein2018functionally}. Additionally, each cell line possesses a unique gene expression profile and signaling pathway activity, leading to varied responses to identical perturbations \cite{way2022morphology}. If a model relies solely on cell line-specific information for prediction, it will fail to generalize to \textit{de novo} cell lines, as such information is not transferable across cell lines. However, completely disregarding cell line-specific information can also lead to suboptimal model performance, particularly when perturbation-specific signals are insufficient for accurate predictions. In such cases, leveraging cell line-specific information can actually enhance generalization performance. The primary challenge lies in deriving useful cell line-specific representations for previously unseen cell lines. To address this, we fine-tune a single-cell foundation model (scFM) \cite{cui2024scgpt,lopez2018deep}  to extract informative transcriptomic features tailored to each cell line. We begin by collecting RNA-seq data for cell lines in both the training set (\(\mathcal{C}^{tr}\)) and the test set (\(\mathcal{C}^{te}\)) from the basal gene expression dataset of human-derived cell lines in the GSE portal \cite{barrett2012ncbi}, selecting \( k \) highly variable genes. Let \( E_c \in \mathbb{R}^k \) denote the gene expression vector for a given cell line \( c \in \mathcal{C}^{tr} \cup \mathcal{C}^{te} \). We then fine-tune a pretrained scFM on a cell-type annotation task, resulting in a refined cell line-specific representation:
\[
h_c = \text{scFM}(E_c).
\]
This transformation effectively converts high-dimensional, sparse RNA-seq data into a compact, information-rich embedding. By incorporating these refined cell line-specific representations, our approach balances the need for generalization while leveraging biologically relevant information to enhance predictive performance.

\subsubsection{Integrating perturbation-specific and cell line-specific Information}

Inspired by the training paradigm of vision-language models, we conceptualize perturbation-specific and cell line-specific information as sequences of tokens and integrate them using a ViT during pretraining. Specifically, given a cell line-specific representation \( h_c \), we first transform it into a sequence of \( m \) transcriptomic tokens, denoted as \( T^c = [t^c_1, \ldots, t^c_m] \), as illustrated in the lower left part of \autoref{fig:2}. These cell line-specific tokens are then prepended as a prefix to the sequence of perturbation-specific image tokens \( T^p = [t^p_1, \ldots, t^p_n] \), forming a unified input for the vision encoding model:
\[
z = \text{ViT}([T^c, T^p]), \quad z = [z_1, z_2, \dots, z_{m+n}].
\]
Here, the image tokens \( z_{m+1}, \dots, z_{m+n} \) are averaged to obtain the final visual representation, which is subsequently used for downstream prediction tasks. This token-based fusion mechanism enables the model to jointly process perturbation-specific and cell line-specific information, enhancing its ability to learn biologically meaningful representations that generalize across diverse cell lines.

\section{Experiment}
\begin{figure}[t]
    \centering
    \includegraphics[width=1\linewidth]{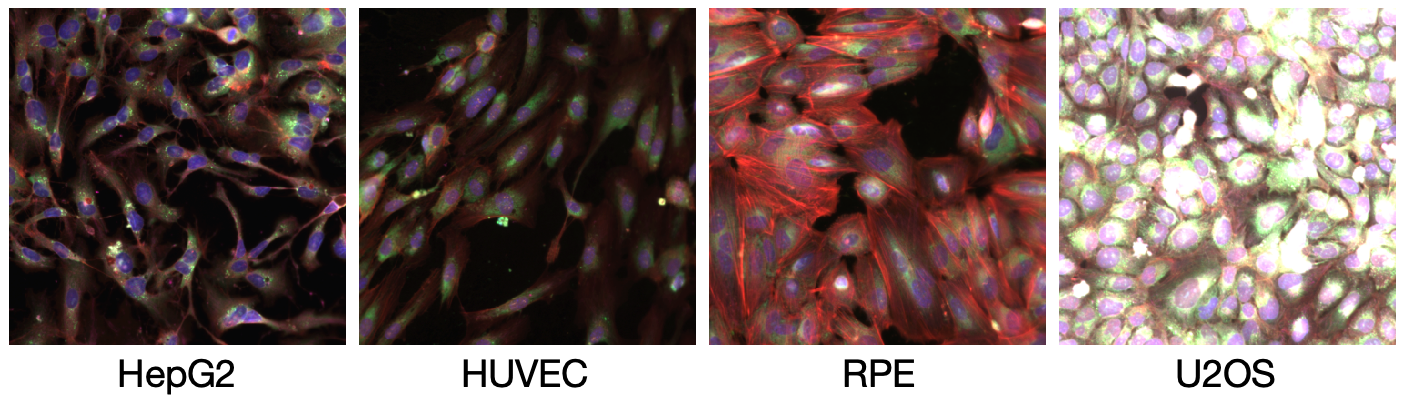}
    \vspace{-18pt}
    \caption{Comparison of cellular phenotypes in response to the same genetic perturbation (gene \textit{GAA}) across different cell lines.}
    \label{fig:3}
    \vspace{-2pt}
\end{figure}
We empirically evaluate the effectiveness of our approach for the perturbation screening task in a \textit{de novo} cell line setting using two large-scale cellular microscopy imaging datasets: RxRx1 \cite{rxrx1} and RxRx19a \cite{cuccarese2020functional}. Our evaluation follows a structured approach: first, we benchmark the performance of existing pretraining methods on microscopy images to establish a baseline. Next, we demonstrate the impact of incorporating our proposed method, which integrates biological knowledge during model pretraining, to enhance robust representation learning. Furthermore, we conduct a systematic analysis to disentangle the contributions of graph regularization loss and cell line-specific representation derived from single-cell foundation models, providing deeper insights into their respective roles in improving model generalization to unseen cell lines.
\subsection{Experimental Setup}
\label{setup}
\begin{table*}[t]
\centering
    \begin{tabular}{l|lcccccc}
    \toprule
\multirow{2}{*}{Pretrain} & Cell Line& \multicolumn{2}{c}{U2OS (RxRx1)}  & \multicolumn{2}{c}{HRCE (RxRx19a)} & \multicolumn{2}{c}{VERO (RxRx19a)}\\
 & Metrics & Top-1 &Top-5&  Top-1 &Top-5&  Top-1 &Top-5 \\
      \midrule
      None & Baseline & $0.09\pm0.02$ & $0.47\pm0.12$ & $0.07\pm0.00$ & $0.33\pm0.01$ & $3.21\pm0.02$ & $16.02\pm0.23$\\
      \midrule
\multirow{2}{*}{{WSL \cite{rxrx1,moshkov2024learning}}}   & baseline &$ 4.12\pm0.10$ & $8.84\pm0.35$  & $3.57\pm0.19$ & $8.73\pm0.42$ & $34.11\pm1.41$ & $72.26\pm2.24$\\
&  Ours\cellcolor{gray!20}  &  $4.79\pm0.16$ \cellcolor{gray!20}&  $9.60\pm0.16$\cellcolor{gray!20} &$4.24\pm0.19$\cellcolor{gray!20} &$9.78\pm0.38$\cellcolor{gray!20} &$38.95\pm2.61$\cellcolor{gray!20}  &  $75.89\pm2.36$ \cellcolor{gray!20}\\

\midrule
\multirow{2}{*}{SimCLR \cite{bendidi2024exploring,perakis2021contrastive}}   & baseline & $4.22\pm0.20$ & $8.57\pm0.43$ & $3.68\pm0.23$& $9.05\pm0.33$& $32.82\pm2.23$ & $72.90\pm2.86$\\

 &  Ours\cellcolor{gray!20} &  $4.59\pm0.22$\cellcolor{gray!20} &   $9.07\pm0.36$\cellcolor{gray!20} &  $3.99\pm0.21$\cellcolor{gray!20} &  $9.21\pm0.20$\cellcolor{gray!20}  &   $38.71\pm1.55$\cellcolor{gray!20} &  $75.00\pm1.61$\cellcolor{gray!20}\\
\midrule
\multirow{2}{*}{BYOL \cite{bendidi2024exploring}}   & baseline & $3.61\pm0.22$ & $8.02\pm0.18$ & $3.39\pm0.36$ & $8.92\pm0.50$ & $31.37\pm4.91$ & $69.92\pm2.37$\\
& Ours\cellcolor{gray!20} & $3.80\pm0.25$\cellcolor{gray!20} & $8.05\pm0.15$\cellcolor{gray!20} & $3.53\pm0.10$\cellcolor{gray!20} & $9.03\pm0.32$\cellcolor{gray!20} & $35.00\pm3.63$\cellcolor{gray!20} & $70.73\pm3.80$\cellcolor{gray!20}\\
\midrule
\multirow{2}{*}{MoCo v3  \cite{chen2021empirical}}   & baseline & $2.18\pm0.15$ & $5.80\pm0.34$ & $2.20\pm0.06$ & $5.97\pm0.29$ &$20.73\pm5.67$ & $53.31\pm9.06$\\
& Ours\cellcolor{gray!20} & $2.56\pm0.10$\cellcolor{gray!20}& $6.36\pm0.20$\cellcolor{gray!20}& $2.53\pm0.17$\cellcolor{gray!20} & $6.86\pm0.29$\cellcolor{gray!20}&$25.56\pm4.27$\cellcolor{gray!20} & $62.26\pm4.99$\cellcolor{gray!20}\\
\midrule
\multirow{2}{*}{MAE \cite{kraus2024masked}}   & baseline & $1.83\pm0.16$ & $5.32\pm0.16$& $1.24\pm0.32$ & $3.95\pm0.57$ & $23.06\pm4.75$ & $58.23\pm7.35$\\
&  Ours\cellcolor{gray!20}  &   $2.10\pm0.10$\cellcolor{gray!20} &   $5.83\pm0.17$\cellcolor{gray!20} &  $1.79\pm0.33$\cellcolor{gray!20}& $5.17\pm0.52$\cellcolor{gray!20} &  $24.60\pm3.05$\cellcolor{gray!20} & \cellcolor{gray!20}$63.31\pm3.39$\\
\bottomrule
    \end{tabular}
    \caption{Perturbation screening performance on novel cell lines after few-shot fine-tuning. We report the mean and standard deviation of top-1 and top-5 accuracies (in percentage) over five random splits for various pretraining approaches, including baseline methods and those augmented with our proposed method.  ``None'' indicates that the ViT was directly initialized with ImageNet-pretrained weights without any additional pretraining on microscopy images. The rows corresponding to our method are highlighted in grey.}
    \label{tab:2}
\end{table*}

\subsubsection{Datasets}
\paragraph{Microscopy Imaging Datasets.}
RxRx1 consists of high-throughput microscopy images capturing a diverse range of genetic perturbations across multiple cell lines. For our experiments, we select the HUVEC, RPE, and HepG2 cell lines for pretraining and fine-tune the model on the U2OS cell line to assess generalization. In contrast, RxRx19a is the first microscopy imaging dataset designed to study the rescue of COVID-19-induced morphological effects, containing images from the HRCE and Vero cell lines under chemical perturbations. Unlike RxRx1, RxRx19a images contain five channels instead of six due to the absence of the MitoTracker stain, which increases the difficulty of transfer learning. To ensure significant perturbation effects, we select only the highest dose samples for each chemical perturbation. \autoref{tab:data} summarizes key dataset statistics, while Appendix B.1 details the preprocessing steps. Additionally, \autoref{fig:3} presents microscopy images illustrating how different cell lines respond to the same genetic perturbations. These visualizations reveal substantial phenotypic variations among cell lines subjected to identical perturbations, highlighting the challenge of generalizing perturbation screening performance across diverse cell lines.
\begin{table}[t]
    \centering
    \resizebox{1.0\linewidth}{!}{
    \begin{tabular}{l|cccc}
    \toprule
     Dataset   & Pretrain  &U2OS (R1) & HRCE (R19) & VERO (R19)\\
     \midrule
   \#  Plate   & 46 & 5 & 53 & 4\\    
  \# Perturbation   & 1138 & 1138 & 1512 & 31\\
\# Image (train)& 101486  &  1138 & 1512 & 62 \\

    \bottomrule
    \end{tabular}} \vspace{-6pt}
    \caption{Summary statistics of the RxRx1 and RxRx19a datasets.}
    \label{tab:data}
\end{table}

\paragraph{Biological Knowledge Sources.}
In our work, we utilize basal gene expression data from 53 human-derived cell lines, characterized using the human whole transcriptome assay (GSE288929)\footnote{Vero cell line is from GSM7745109.}, to obtain robust cell line-specific representations. To achieve this, we explore two single-cell foundation models—scGPT \cite{cui2024scgpt} and scVI \cite{lopez2018deep}—both of which are fine-tuned on a cell type annotation task to generate informative embeddings that effectively capture cell line-specific characteristics. For constructing the perturbation relational graph, we integrate external gene-gene interaction knowledge from two major biological databases: STRING and Hetionet. From the STRING database, we select human data with the network type set to "combined score", applying a confidence threshold of 200 to filter reliable gene-gene interactions with quantitative scores. In contrast, Hetionet provides binary connections between genes, for which we perform a random walk to derive a probability matrix that reflects interaction likelihoods. Additionally, inspired by \citet{liu2025learning}, we compute gene-gene similarities based on raw image features extracted from the dataset\footnote{https://www.rxrx.ai/rxrx1\#Download}. To construct a comprehensive and biologically meaningful gene interaction weight matrix, we combine and normalize these three sources—STRING, Hetionet, and image-derived similarities. For drug perturbations in RxRx19a, we utilize the STITCH database \cite{szklarczyk2016stitch}, which directly provides drug-drug interactions. This integration of multi-source biological knowledge ensures that our model captures meaningful perturbation relationships and enhances its generalization to unseen cell lines.

\begin{table*}[t]
\centering
    \begin{tabular}{cc|cccccccccc}
    \toprule
  \multicolumn{2}{c}{Method}&   \multicolumn{2}{c}{WSL }& \multicolumn{2}{c}{SimCLR} & \multicolumn{2}{c}{BYOL}& \multicolumn{2}{c}{MoCo v3  } & \multicolumn{2}{c}{MAE}  \\
 w. CS& w. PS & Top-1 &Top-5&  Top-1 &Top-5&  Top-1 &Top-5&  Top-1 &Top-5& Top-1 &Top-5  \\
      \midrule
 $\times$  & $\times$ & 4.12 & 8.84 & 4.22 & 8.57 & 3.61& 8.02 & 2.18  & 5.80 & 1.83 & 5.32\\
$\checkmark$& $\times$&  4.64 &9.48 & 4.39 & 8.76 &  3.71 & 7.95 &  2.45 &  6.21&1.95&5.42\\
$\times$ & $\checkmark$ & 4.25 & 9.09 & 4.48 &  8.90 &\textbf{3.97} &\textbf{8.35}  & 2.33  & 6.08 &1.98&5.61\\
$\checkmark$&$\checkmark$ & \textbf{4.79} & \textbf{9.60}& \textbf{4.59} & \textbf{9.07} & {3.80} & {8.05} & \textbf{2.56} & \textbf{6.36}  &\textbf{2.10} & \textbf{5.83}\\
\bottomrule
    \end{tabular}
    \caption{Ablation study on the U2OS cell line. This table reports the average top-1 and top-5 accuracies over five random splits. The best results are highlighted in bold. CS and PS denote cell line-specific representation and perturbation-specific representation, respectively.}
    \label{tab:ab}
\end{table*}

\subsubsection{Baselines}
\label{baseline}
We conduct experiments using existing pretraining methods designed for microscopy cell images, including weakly supervised learning (WSL) \cite{zhou2018brief} and various self-supervised learning approaches such as SimCLR \cite{chen2020simple}, BYOL \cite{grill2020bootstrap}, MoCo v3 \cite{chen2021empirical}, and MAE \cite{he2022masked}. Given the scaling properties of microscopy image learning discussed in \citet{kraus2024masked} and the size constraints of our pretraining dataset, we select ViT-S/16 as the vision encoding model for all our experiments. To obtain the embeddings used in our graph regularization loss, we utilize the main encoder of WSL, SimCLR, and MAE. For methods that involve multiple vision encoders, such as BYOL and MoCo v3, we extract embeddings from the student encoder.
\subsubsection{Evaluation setting}
We begin by pretraining the vision encoder using baseline methods on the training set of the RxRx1 dataset. To assess generalization in a de novo cell line setting, we then perform one-shot supervised classification fine-tuning on the testing set of RxRx1 and few-shot fine-tuning on the RxRx19a dataset, measuring classification accuracy. Experiments are conducted across five random splits, and we report the average top-1 and top-5 accuracy to ensure robustness. Beyond evaluating existing pretraining methods, we further enhance the models by integrating our proposed components—the perturbation relational graph regularization loss and cell line-specific representations derived from scFM. The same experimental pipeline is applied across both standard and enhanced model settings. All experiments are conducted on a machine equipped with eight NVIDIA A100 GPUs. Additional details regarding pretraining and fine-tuning configurations are provided in Appendix B.5.

\subsection{Main Results}
\autoref{tab:2} presents the perturbation screening performance on three de novo cell lines—U2OS from RxRx1 and HRCE and Vero from RxRx19a—after few-shot fine-tuning. We compare several existing pretraining methods as baselines and evaluate the impact of incorporating biological knowledge through our proposed approach. Our results highlight two key observations. First, integrating biological knowledge significantly improves model performance across all baseline pretraining methods, confirming its role in enhancing generalization to novel cell lines. In particular, for the U2OS cell line, our method achieves an average top-1 accuracy improvement of approximately 20\% over the baseline, with notable gains in both top-1 and top-5 accuracy for the other two cell lines as well. Second, we observe that ViT models pretrained using different methods exhibit a substantial performance boost compared to unpretrained models, suggesting that pretraining on different cell lines still provides generalizable features useful for novel cell lines. Among the baseline pretraining approaches, WSL outperforms self-supervised techniques, likely due to its direct optimization for perturbation classification, which aligns more closely with our downstream task. However, under one-shot fine-tuning, we observe a accuracy drop for HRCE cell line compared to U2OS. This discrepancy is likely due to differences in imaging protocols between RxRx1 and RxRx19a, making model adaptation more challenging for HRCE. These findings emphasize the importance of developing pretraining strategies that can robustly adapt to de novo cell lines while accounting for variability in experimental conditions.

\subsection{Ablation Studies}
\subsubsection{Analysis of perturbation-specific and cell line-specific representations}
\label{ab}

In this section, we perform an ablation study on the U2OS cell line to assess the individual contributions of perturbation-specific and cell line-specific representations. The results, presented in \autoref{tab:ab}, demonstrate that both representations lead to significant improvements over the baseline, with their combined use typically achieving the best performance. Notably, the improvement from perturbation-specific representations is more pronounced on SimCLR, BYOL and MAE than that from cell line-specific representations. This discrepancy may stem from the limited diversity of our pretraining dataset, which consists of only three cell lines, whereas the actual heterogeneity among cell lines is far more complex. Consequently, cell line-specific representations may be prone to overfitting, limiting their effectiveness. Additionally, we observe that adding cell line-specific representations to BYOL results in a performance drop.  In contrast, the inclusion of the perturbation relational graph consistently leads to substantial accuracy gains, both when used in isolation and in combination with cell line-specific representations. These findings validate the effectiveness of our proposed approach, highlighting the importance of integrating external biological knowledge for improved generalization in \textit{de novo} cell line settings.

\begin{figure}[t]
  \centering
  \begin{subfigure}{0.495\linewidth}
    \includegraphics[width=\linewidth]{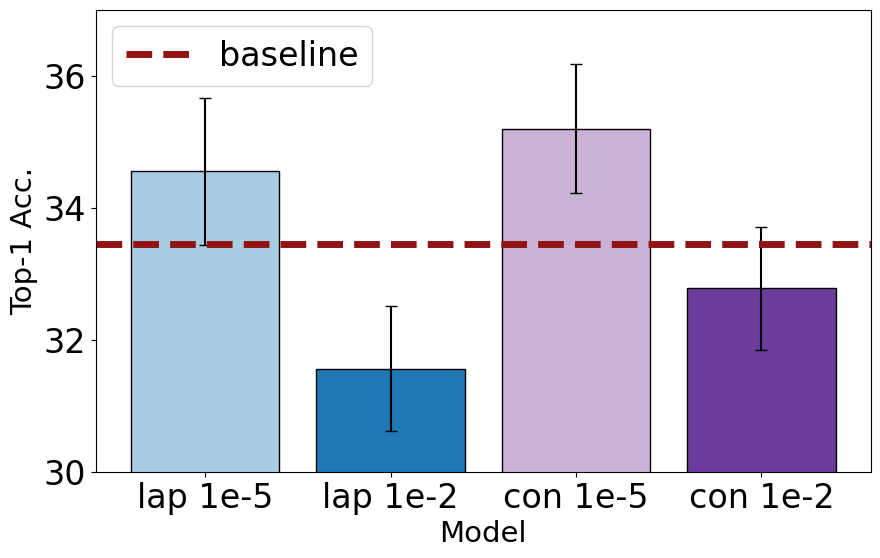}
    \caption{Split 1: same batch.}
    \label{fig:sp1}
  \end{subfigure}
  \begin{subfigure}{0.495\linewidth}
    \includegraphics[width=\linewidth]{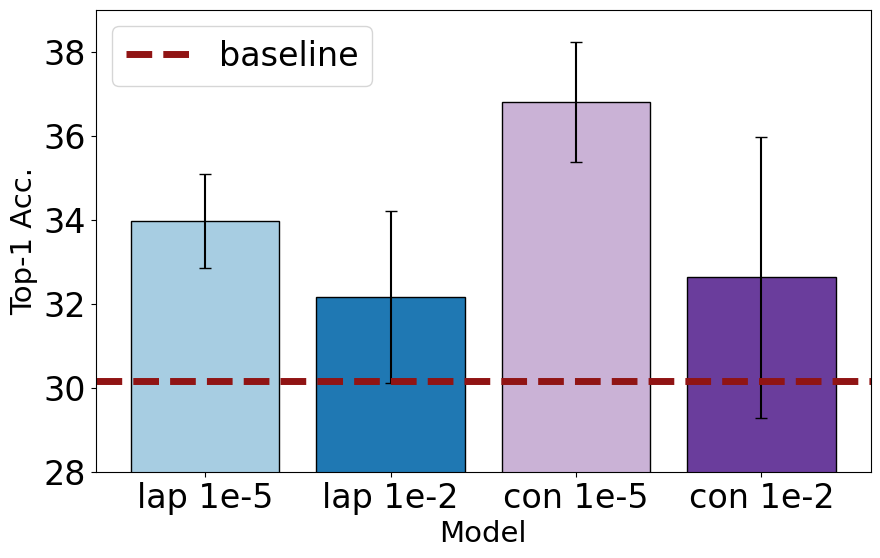}
    \caption{Split 2: different batches.}
    \label{fig:sp2}
  \end{subfigure}
  \caption{Analysis of graph regularization losses on the VERO dataset under different data split conditions for 2-shot fine-tuning. For each split condition, we generate five random splits and report the average top-1 accuracy.  ``lap'' and ``con'' refer to the graph Laplacian loss and graph node contrastive loss, respectively, and 1e-2 and 1e-5 represent the corresponding loss weights.}
  \label{fig:ab1}
\end{figure}

\label{analysis}
\subsubsection{Analysis of graph regularization losses}
\label{ab:loss}
To evaluate graph regularization losses, we used a ViT model pretrained with SimCLR and performed 2-shot fine-tuning on the VERO cell line. We selected the SimCLR-pretrained ViT because the addition of graph loss led to significant performance gains over the baseline while also exhibiting high variance, making it ideal for analyzing the impact of graph losses and their interaction with batch effects in the training data. To systematically study these effects, we partitioned the fine-tuning data into two distinct split conditions: (1) In split condition 1, training samples were selected from \textit{a single plate}, while test samples came from the remaining plates within the same batch. (2) In split condition 2, training samples were chosen across \textit{multiple batches}, with test samples consisting of the remaining dataset (excluding the training plates). \autoref{fig:ab1} presents the results.

We observe that in split condition 2, incorporating graph regularization loss significantly improves performance compared to the baseline. Notably, the graph contrastive loss yields the most pronounced improvement when its weight is set to 1e-5, and overall, using a 1e-5 weight for both graph contrastive ($\mathcal{L}_{con}$) and graph Laplacian ($\mathcal{L}_{lap}$) losses results in better performance than using 1e-2. In contrast, in split condition 1, the impact of regularization loss is relatively minimal, likely because training samples originate from different sites within the same plate, leading to low intra-class variability. However, in split conditions 2, where training samples come from different plates and batches, the greater intra-class variability makes graph regularization loss more effective, further validating Proposition 3.1. These findings suggest that in practical applications, where data typically span multiple plates and batches, incorporating graph regularization losses is particularly beneficial for enhancing model robustness and generalization.  More detailed explanations and analyses are provided in the Appendix C.1.

\begin{figure}
  \centering
  \begin{subfigure}{0.495\linewidth}
    \includegraphics[width=\linewidth]{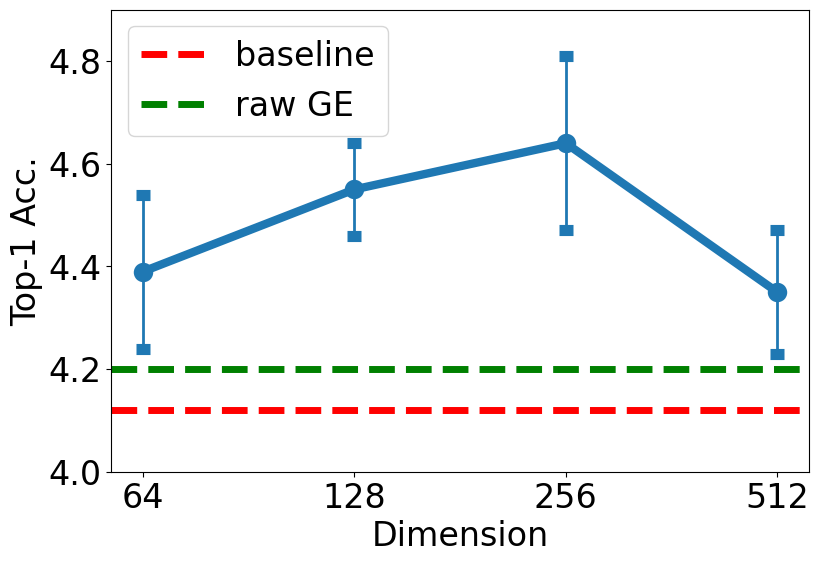}
    \caption{scVI}
    \label{fig:2sp1}
  \end{subfigure}
  \begin{subfigure}{0.495
  \linewidth}
    \includegraphics[width=\linewidth]{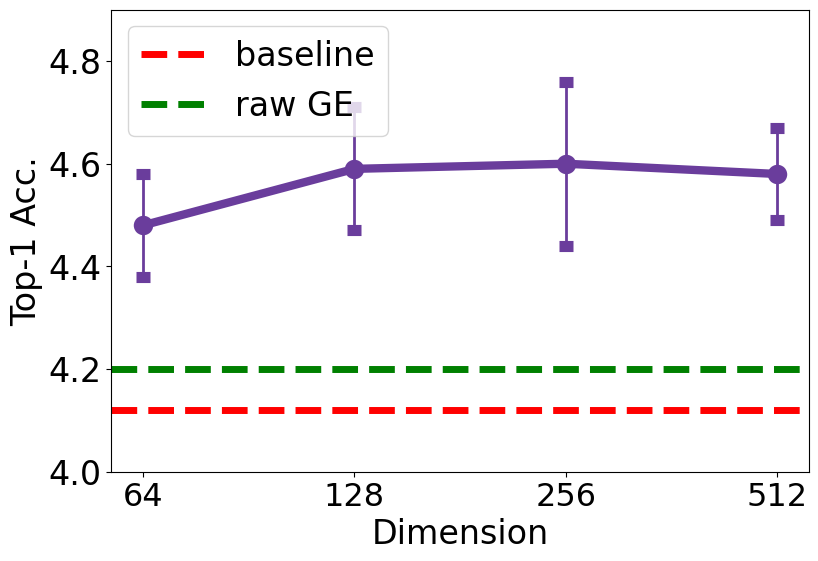}
    \caption{scGPT}
    \label{fig:2sp2}
  \end{subfigure}
  \caption{Analysis of scVI and scGPT on the effect of representation dimensionality. The figure shows the mean and standard deviation of top-1 accuracy, with ``Raw GE'' indicating results using raw cell line gene expression for comparison.}
  \label{fig:ab2}
\end{figure}

\subsubsection{Analysis of different scFMs}
To assess the robustness of cell line-specific representations, we conducted experiments exploring different scFMs and varying the output gene expression token dimensionality. Using WSL as the pretraining method, we evaluated model performance on the U2OS cell line, with results presented in \autoref{fig:ab2}. Our findings indicate that both scVI and scGPT models achieve the highest accuracy when the representation dimensionality is set to 128 or 256, whereas increasing the dimensionality to 512 leads to a noticeable performance decline. This suggests that overly high-dimensional representations may introduce redundant or less informative features, reducing their effectiveness for downstream tasks. Additionally, when using raw gene expression data instead of learned representations, we observe a slight improvement over the baseline, but the performance gains remain less pronounced compared to the representations generated by scFMs. These results highlight the importance of leveraging scFMs to extract compact and informative cell line-specific embeddings, ultimately enhancing model generalization to \textit{de novo} cell lines. Notably, scGPT, which was trained on extensive single cell data, did not demonstrate significant advantages over scVI, a finding that aligns with similar observations reported in previous studies \cite{csendes2025benchmarking}.

\section{Conclusion}
In this paper, we tackle the challenge of developing robust microscopy image profiling models for novel cell lines. We introduce an approach that integrates external biological knowledge to guide models in learning both perturbation-specific and cell line-specific representations, enabling improved generalization to unseen cell lines. Our method is evaluated on two RxRx datasets, where few-shot fine-tuning on multiple novel cell lines demonstrates significant improvements in perturbation screening performance. These results highlight the effectiveness of incorporating biological priors into model pretraining to enhance transferability. For future work, we aim to extend our approach to larger perturbation datasets and explore its applicability to additional tasks in novel cell lines, such as mechanism-of-action classification, further expanding its impact in drug discovery.

\section*{Acknowledgments}
We acknowledge the funding for the project provided by the National Institute of General Medical Sciences (R01GM141279), the National Institute of Allergy and Infectious Diseases (R01AI188576), and the National Science Foundation (2145625).

{
    \small
    \bibliographystyle{ieeenat_fullname}
    \bibliography{main}
}
\newpage



\appendix

\section{Proof of Proposition 3.1}
\label{proof}
Let $\phi: \mathcal{X} \to \mathbb{R}^d$ be a vision encoding model mapping each cellular microscopy image $x \in \mathcal{X}$ to a feature vector. For a given perturbation $v \in \mathcal{V}$, denote by $\mathcal{X}_v \subset \mathcal{X}$ the set of replicate images corresponding to v and define the perturbation node feature as: $f_v = \frac{1}{|\mathcal{X}v|} \sum_{x \in \mathcal{X}v} \phi(x).$
Consider a perturbation relational graph where each node $v \in \mathcal{V}$ is assigned the feature $\psi(v):= f_v$. Then, minimizing the graph regularization loss:
$$\mathcal{L}_{\mathrm{reg}} = \operatorname{tr}\left(F^\top (D - W) F\right),$$
implicitly minimizes the intra-class distance:
$$\frac{2}{|\mathcal{X}_v|\, (|\mathcal{X}_v|-1)} \sum_{x, x{\prime} \in \mathcal{X}_v} \|\phi(x) - \phi(x{\prime})\|.$$
\textit{Proof.}

We first review the definitions that will be used in this proof. 
The node feature matrix is defined as: $$
F = [ f_{v_1}, f_{v_2},\cdots,f_{v_{|\mathcal{V}|}} ]^{\top},
 f_v = \frac{1}{|\mathcal{X}_v|} \sum_{x \in \mathcal{X}v} \phi(x).$$
The adjacency matrix $W$ have elements $w_{uv}$ representing the weight between nodes $u$ and $v$, and the degree matrix $D$ is defined by $
D_{uu} = \sum_{v \in \mathcal{V}} w_{uv}
$. 

Expanding the graph Laplacian loss, we have:
\begin{align*}
&\operatorname{tr}(F^T (D-W) F) = \operatorname{tr}(F^T D F) - \operatorname{tr}(F^T W F)\\
  = &\sum_{u \in \mathcal{V}} D_{uu} \|f_u\|^2 -  \sum_{u,v \in \mathcal{V}} w_{uv} \langle f_u, f_v \rangle.
\end{align*}
By substituting the node feature definition into the inner product. For any two nodes $u,v \in \mathcal{V}$,
$$
\langle f_u, f_v \rangle = \frac{1}{|\mathcal{X}_u|\, |\mathcal{X}_v|} \sum_{x \in \mathcal{X}_u} \sum_{x{\prime} \in \mathcal{X}_v} \langle \phi(x), \phi(x{\prime}) \rangle.
$$
Therefore, when minimizing the graph Laplacian loss, the second term in the loss will be maximized. So given a graph with a fixed adjacency matrix, minimizing the graph Laplacian loss will  maximize:
$$
\frac{1}{|\mathcal{X}_u|\, |\mathcal{X}_v|} \sum_{x \in \mathcal{X}_u} \sum_{x{\prime} \in \mathcal{X}_v} \langle \phi(x), \phi(x{\prime}) .\rangle
$$
In particular, for a fixed perturbation $v$, the average inner product is maximized when all the feature vectors $\phi(x)$ for $x \in \mathcal{X}_v$ are identical. That is, when:
$$
\forall x \in \mathcal{X}_v,\quad \phi(x) = f_v.
$$
This conclusion follows from the optimality properties of the arithmetic mean in vector spaces: the arithmetic mean minimizes the sum of squared distances, thereby maximizing the pairwise inner products under fixed energy.

Since maximizing the inner product term is equivalent to minimizing the pairwise distance between features, minimizing the graph regularization loss implicitly minimizes the intra-class distance:
$$\frac{2}{|\mathcal{X}_v|\, (|\mathcal{X}_v| - 1)} \sum_{x,x{\prime} \in \mathcal{X}_v} \|\phi(x) - \phi(x{\prime})\|.$$
\section{Implementation Details}
\subsection{RxRx Dataset Pre-processing}
\label{preprocess}
For the RxRx1 dataset, the images are originally 512×512 pixels, while the RxRx19a images are 1024×1024 pixels. During training, we perform a random crop to obtain 256×256 patches, and for validation and testing, we use center crops. The RxRx1 dataset contains six channels: Hoechst, ConA, Phalloidin, Syto14, MitoTracker, and WGA. In contrast, RxRx19a lacks the MitoTracker channel; therefore, during weight transfer, we retain the pre-trained model weights corresponding to the channel order in RxRx19a. Additionally, RxRx19a includes both infectious disease and small molecule perturbations. We filter the dataset to retain only the small molecule perturbations, and for each drug treatment, we select the sample corresponding to the highest concentration to ensure significant perturbation effects.

RxRx1 includes four cell lines—HUVEC, RPE, U2OS, and HepG2—while RxRx19a comprises two cell lines, HRCE and VERO, with the number of images for each cell line summarized in \autoref{tab:1}. To ensure that our model learns robust cell line- specific and perturbation-specific representations, we use HUVEC, RPE, and HepG2 in RxRx1 dataset for pretraining. Unlike the split used by \citet{koh2021wilds}, for fine-tuning on U2OS cell line, the training, validation, and test sets are randomly selected from all available plates.
\begin{table}[h]
    \centering
    \resizebox{1.0\linewidth}{!}{
    \begin{tabular}{c|ccccccc}
    \toprule
       Cell Line  & HUVEC &  HepG & RPE  & U2OS & VERO & HRCE\\
       \midrule
      \# Image   &58,848 &26,972 &26,972 & 12,260&1,984 &36,896\\
      \bottomrule
    \end{tabular}}
    \caption{Summary of cell line image data in the RxRx1 and RxRx19a datasets.}
    \label{tab:1}
\end{table}
\subsection{External Biological Knowledge}
In our work, we leverage three external databases to enhance the connectivity between perturbations, guiding the model to learn perturbation-specific representation. The STRING database \cite{szklarczyk2023string} aggregates information on gene interactions—including physical protein-protein interactions, co-expression patterns, and shared functional pathways—and provides quantitative confidence scores that serve as edge weights, reflecting the strength of gene-gene interactions. To further reinforce the relationships between perturbations, we incorporate Hetionet \cite{himmelstein2017systematic}, which offers curated gene-gene associations in a binary format. We apply a 5-step random walk on the Hetionet data to capture multi-hop information, thereby enriching the connectivity beyond direct interactions. Finally, we integrate these sources with similarity scores computed from the dataset’s raw image features, yielding a comprehensive perturbation relational graph based on gene interactions. For the RxRx1 siRNA perturbations, we query ThermoFisher’s database to retrieve their target genes. In cases where an siRNA corresponds to multiple genes (e.g., s19148 and s27703), we retain only the first gene from the query results. Using this mapping, we observe that siRNAs s18582 and s18583 both target the gene IQCB1, which explains why their post-perturbation images are very similar. We obtain these gene-gene interactions by calling the STRING and Hetionet APIs.

For RxRx19a with drug perturbations, we directly query the STITCH database\footnote{\url{http://stitch.embl.de/download/chemical_chemical.links.v5.0.tsv.gz}} to obtain chemical–chemical relationships. STITCH essentially establishes these relationships by leveraging the target proteins of drugs, as illustrated in \autoref{fig:ab2}.

\begin{figure}[t]
  \centering
  \begin{subfigure}{0.48\linewidth}
    \includegraphics[width=\linewidth]{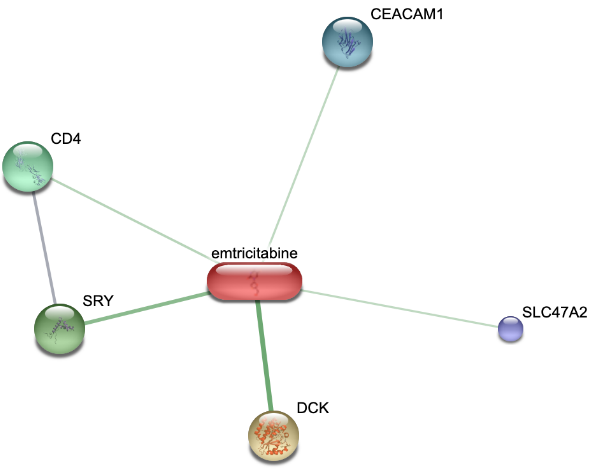}
    \caption{emtricitabine}
    \label{fig:2sp1}
  \end{subfigure}
  \begin{subfigure}{0.48
  \linewidth}
    \includegraphics[width=\linewidth]{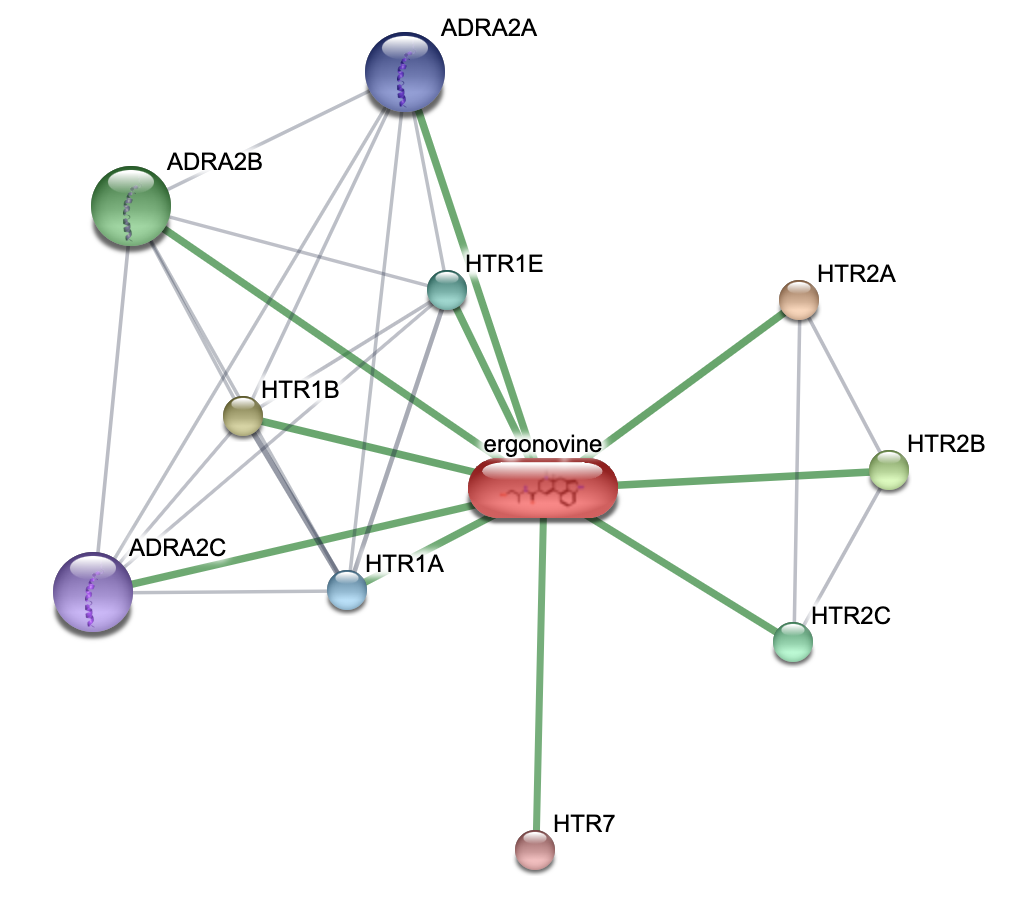}
    \caption{ergonovine}
    \label{fig:2sp2}
  \end{subfigure}
  \caption{Illustration of drug–gene relationships, where drug target proteins serve as the linking mechanism between different chemical perturbations}
  \label{fig:ab2}
\end{figure}

Cell line RNA sequencing (RNA-seq) is a high-throughput technique that provides a comprehensive snapshot of gene expression profiles in cultured human cells. This technology quantifies transcript abundance for thousands of genes simultaneously, offering valuable insights into the molecular state and functional pathways active in different cell lines. In our work, we obtained raw gene expression data for human cell lines from Gene Expression Omnibus GSE288929 and GSM7745109, where over 26,000 genes were sequenced. Our preprocessing pipeline uses Scanpy \cite{virshup2023scverse} to filter out genes expressed in fewer than 5 cell lines, followed by total count normalization with a target sum of 1e4 and a log1p transformation. Finally, we identify 1,000 highly variable genes using Scanpy, which serve as the input for our single-cell foundation models.

\subsection{Single-cell Foundation Models}
Single-cell foundation models (scFMs) are robust pre-trained models designed to integrate and analyze the complex, multimodal data generated by single-cell sequencing technologies, enabling tasks such as cell-type annotation and gene function prediction. Inspired by the success of large language models in NLP, these scFMs leverage extensive single-cell datasets to capture intricate intra- and inter-cellular relationships, although their development and comprehensive evaluation remain in the early stages compared to other fields. scFM can extract informative transcriptomic representations that capture cell line-specific molecular characteristics, which helps the vision model better distinguish perturbation effects from inherent cell line differences. To explore the impact of scFMs, we select two representative models: scVI \cite{lopez2018deep} and scGPT \cite{cui2024scgpt}. scVI (single-cell Variational Inference) is a deep generative model based on variational autoencoders that learns low-dimensional latent representations of cells while accounting for technical noise and batch effects, making it particularly effective for data integration and uncertainty quantification. scGPT is a generative pretrained transformer built on over 33 million cells, which effectively distills critical biological insights through self-attention mechanisms to support diverse downstream tasks in single-cell biology. Additionally, we also explored scFoundation \cite{hao2024large}, a large-scale model with 100 million parameters trained on over 50 million human single-cell transcriptomic profiles, to further validate our findings.

Cell type annotation is the task of assigning each cell a label that reflects its biological identity, based on its gene expression profile. For scVI, we directly train the model using the scanpy implementation with default parameters to obtain cell representations. For scGPT, we follow the provided cell type annotation tutorial\footnote{\url{https://scgpt.readthedocs.io/en/latest/tutorial_annotation.html}} and fine-tune the model on our data for 5 epochs to obtain 512-dimensional features. For representations with dimensions lower than 512, we add a linear projection layer after the [CLS] token to reduce the dimensionality to the desired size.
\subsection{Baselines}
\begin{enumerate}
    \item Weakly Supervised Learning (WSL) approach leverages perturbation labels by assigning the same label to images from the same perturbation, regardless of cell line or experimental batch, thus directly optimizing for perturbation classification.
    \item SimCLR is a self-supervised learning framework that learns robust features by contrasting multiple augmented views of the same image. In our experiments, considering the characteristics of microscopy images, we apply transformations such as random resized cropping, horizontal flipping, rotation, and Gaussian blurring. The projector in SimCLR is implemented as a two-layer MLP with a hidden dimension of 2048 and an output feature dimension of 128, and a temperature of 0.1 is used during contrastive learning.
    \item BYOL (Bootstrap Your Own Latent) employs a teacher-student architecture to learn representations without the need for negative samples. Similar to SimCLR, it uses the same set of augmentations for generating different views, and the output dimension of its projector is set to 2048.
    \item MoCo v3 (Momentum Contrast v3) is a self-supervised method that combines momentum encoding with Vision Transformers for learning visual representations. It employs a dual-encoder architecture where the key encoder is updated via exponential moving average of the query encoder. For our microscopy images, we apply the same set of augmentations as SimCLR and BYOL. The projector and predictor in MoCo v3 are MLPs with hidden dimensions of 2048 and output dimensions of 256. The momentum coefficient starts at 0.99 and increases to 1.0 following a cosine schedule, and a temperature of 0.1 is used for the InfoNCE loss.
    \item MAE (Masked Autoencoders) reconstructs masked portions of the input image, encouraging the model to learn contextual and spatial representations. In our implementation, the mask ratio is set to 0.75, and the decoder is based on the ViT-Small architecture.
\end{enumerate}
In this paper, we explore the role of biological knowledge in enhancing these pre-training methods; therefore, we adopt the general set of hyperparameters for each pre-training method respectively and ensure consistency in their settings when comparing models with and without the integration of our proposed modules.
\subsection{Pre-training and Fine-tuning}
\label{setting}
For all pretraining methods, we adopt ViT-S/16 with class token as the image encoder, utilizing mean pooling for feature aggregation. The same backbone is used for fine-tuning in the perturbation screening task. And for both pretraining and fine-tuning, we apply Typical Variation Normalization (TVN) to the input images to mitigate the influence of batch effects. Following \citet{kraus2024masked}, we use AdamW as the optimizer with a learning rate of 1e-3 and a weight decay of 0.05, and we employ a OneCycleLR scheduler with a cosine annealing schedule. Our training is distributed via DDP with a per-GPU batch size of 400, yielding a global batch size of 3200. We use automatic mixed precision during the pretraining. After extensive experimentation, we set the maximum epoch to 400 for SimCLR, BYOL, MoCo v3 and MAE approaches, while for WSL, we train for 200 epochs. For both the baseline models and those augmented with our method, we adopt identical pretraining settings. In the experiments reported in Table 1 (main paper), to ensure generality, we use scVI as the single-cell foundation model with the output feature dimension of 256. Moreover, we select the graph contrastive loss as our perturbation graph regularization loss. The detailed hyper-parameter settings, such as the number of gene tokens and weight of graph regularization loss, are provided in the github repository. 

During fine-tuning, we load the pretrained ViT and perform few-shot supervised learning on unseen cell lines. The fine-tuning configuration is as follows: a maximum of 200 epochs, AdamW optimizer with a learning rate of 1e-3, and we choose the model with best performance on validation set for the evaluation. Through extensive experiments, we found that adding graph regularization loss during the fine-tuning phase can achieve similar improvements, and is more efficient compared to adding it during the pre-training. Therefore, in our actual implementation, we directly add graph loss during the downstream evaluation.

\begin{figure}[t]
    \centering
    \includegraphics[width=0.98\linewidth]{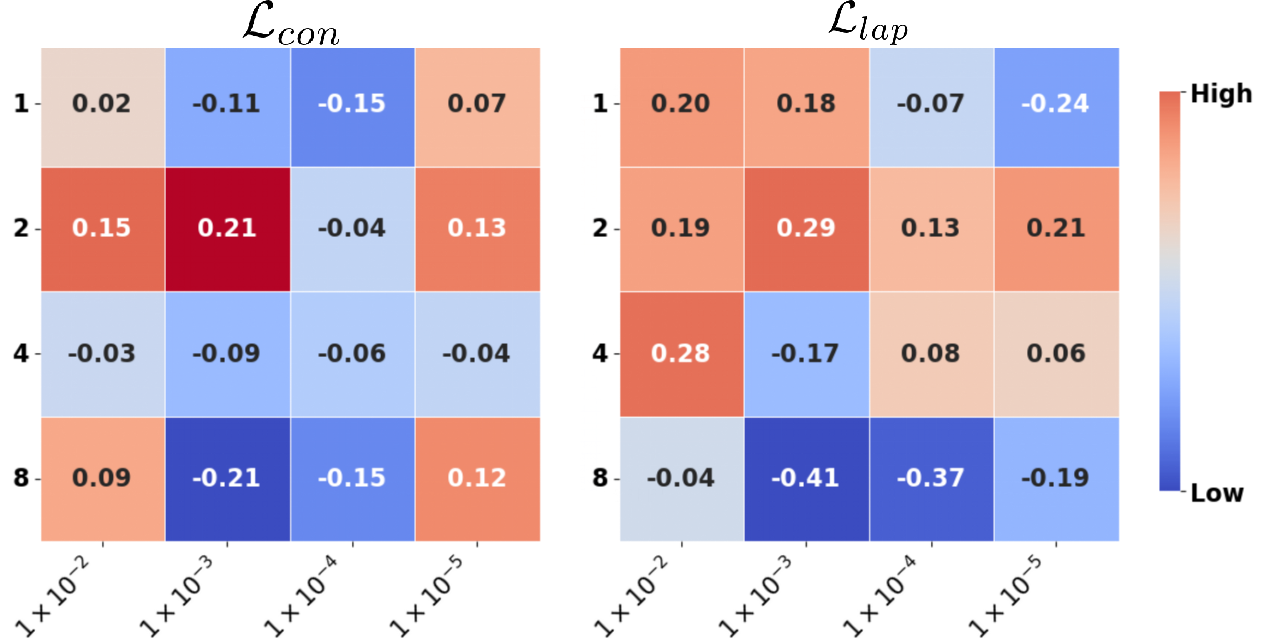}
    \caption{Heatmap of SimCLR experimental results on U2OS cell line compared to the baseline, where the vertical and horizontal axes represent the number of gene expression tokens and the weight of graph regularization loss, respectively. The left heatmap shows the graph contrastive learning loss, and the right shows the graph Laplacian loss.}
    \label{fig:ap1}
\end{figure}

\section{Additional Results}
In this section, we provide additional analyses of the proposed modules, offering deeper insights into their contributions and interactions under various experimental conditions.
\subsection{Analysis of Graph Regularization Losses}

We explore the impact of graph regularization loss and the number of gene expression tokens using SimCLR pre-trained ViT, with results shown in \autoref{fig:ap1}. The gene expression here comes from 128-dimensional representations from scVI. From the figure, we can observe that compared to graph contrastive loss, the graph Laplacian loss generally performs better overall. Additionally, the model performs relatively well when the loss function weight is 1e-2 or 1e-5, but overall performance does not show a linear relationship with weight magnitude. On the other hand, we can find that the model achieves better results when the number of tokens is 2, while performance is less ideal when the number of tokens is 4 or 8, even performing worse than the baseline.

\subsection{Analysis of Gene Expression Tokens}
\begin{figure}[t]
    \centering
    \includegraphics[width=0.98\linewidth]{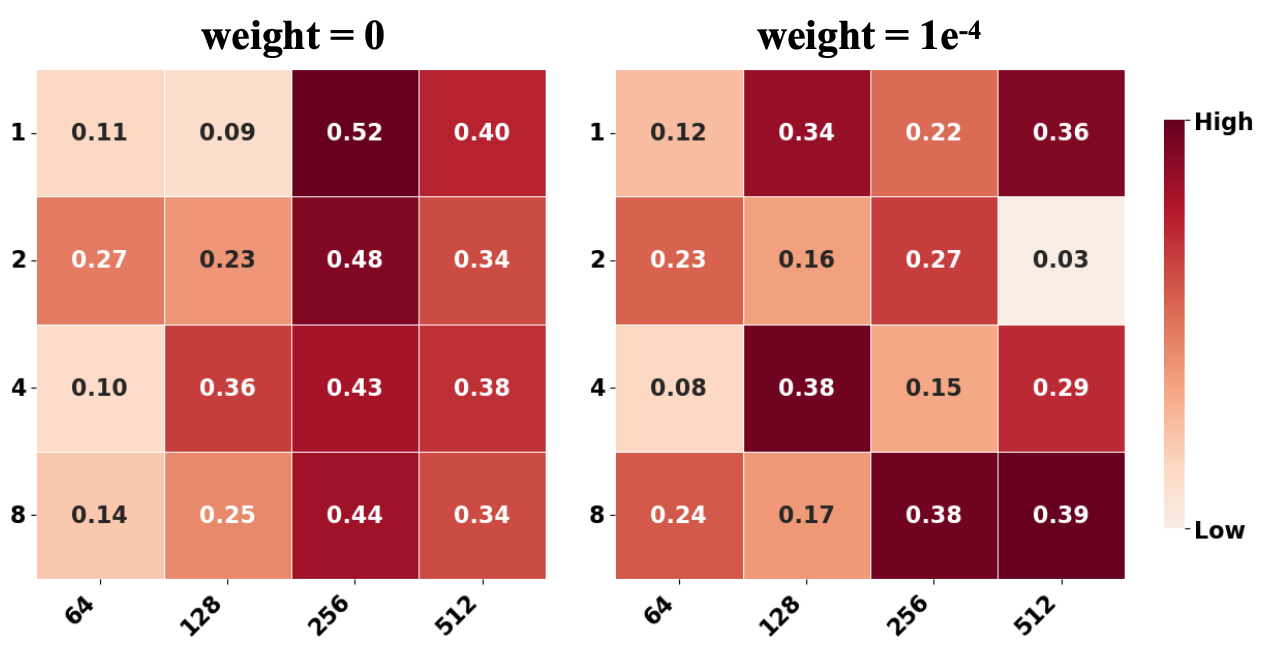}
    \caption{Heatmap of WSL model results under different numbers of gene expression tokens and dimensions, where the tokens are derived from scGPT. The left and right plots represent different graph contrastive learning weights, respectively.}
    \label{fig:ap2}
\end{figure}

We explored the impact of different token numbers and dimensions on WSL pre-trained ViT on U2OS. The results are shown in \autoref{fig:ap2}.We can observe that higher-dimensional tokens, such as 256 and 512, perform relatively better, which may be due to their closer alignment with scGPT's original token dimensions. Additionally, almost all settings show significant improvement over the baseline, regardless of whether graph regularization loss is applied.

\end{document}